\documentclass[journal]{IEEEtran}

\ifCLASSINFOpdf

\else

\fi
\usepackage{amsmath}
\usepackage{graphicx}  
\usepackage{float}
\begin{document}

\title{Guidance Module Network for Video Captioning}

\author{Xiao Zhang\textsuperscript{},
	Chunsheng Liu\textsuperscript{*},
	Faliang Chang\textsuperscript{*}}

\maketitle

\begin{abstract}
Video captioning has been a challenging and significant task that describes the content of a video clip in a single sentence. The model of video captioning is usually an encoder-decoder. We find that the normalization of extracted video features can improve the final performance of video captioning. Encoder-decoder model is usually trained using teacher-enforced strategies to make the prediction probability of each word close to a 0-1 distribution and ignore other words. In this paper, we present a novel architecture which introduces a guidance module to encourage the encoder-decoder model to generate words related to the past and future words in a caption. Based on the normalization and guidance module, guidance module net (GMNet) is built. Experimental results on commonly used dataset MSVD show that proposed GMNet can improve the performance of the encoder-decoder model on video captioning tasks.
\end{abstract}

\begin{IEEEkeywords}
	Video captioning, Normalization of features, Guidance module
\end{IEEEkeywords}

\IEEEpeerreviewmaketitle

\section{Introduction}

\IEEEPARstart{V}{ideo} captioning is widely concerned and it provides a natural language description according to the video content. The development of video captioning can be applied for subsequent video retrieval or summary generation to help visually impaired people understand reality. Video captioning is related to both computer vision and language processing and it is profoundly challenging to describe a video in natural language because of the richness of the content.

Earlier methods \cite{5, 1} define the template and then fill the template with the detected subject, predicate and object, which makes the sentence rigid. In contrast, inspired by the development of neural machine translation (NMT), the sequential learning methods leverage the sequential learning model to translate video contents directly into sentences. The encoder-decoder model is extensively adopted in natural language processing (NLP), and also performs decently in video captioning \cite{N5}. The encoder in general encoder-decoder model is based on recurrent neural network (RNN). Different from NLP, for video captioning, the convolutional neural network (CNN) reads the video and generates the video feature vectors. The representing features are fed into the decoder which is RNN-based model to directly generate the natural language description of the video.

It has been widely accepted that normalization of features in neural networks can improve network performance, but no article in video captioning task explicitly proposes the use of normalization for features. We attempt to apply normalization for video captioning and achieve better performance.

There exists a disadvantage in encoder-decoder framework although it performs excellently in video captioning tasks. Teacher-enforced strategy is commonly applied to train encoder-decoder models to make the prediction probability of each word close to a 0-1 distribution, which causes the network ignores context words. This weakness impinges on the consistency of sentences in translation, which is disadvantaged for describing video content. Aiming at this problem, we propose a guidance module which can make up for this shortcoming. The contributions of this article are as follows:

$\bullet$ We dicover that normalization of input features and features obtained by the soft attention module can improve the performance of the original network for video captioning.

$\bullet$ A guidance module is proposed into the encoder-decoder framwork, which can improve the performance compared to the original encoder-decoder model.

\begin{figure}  
	\centering  
	\includegraphics[scale=0.7]{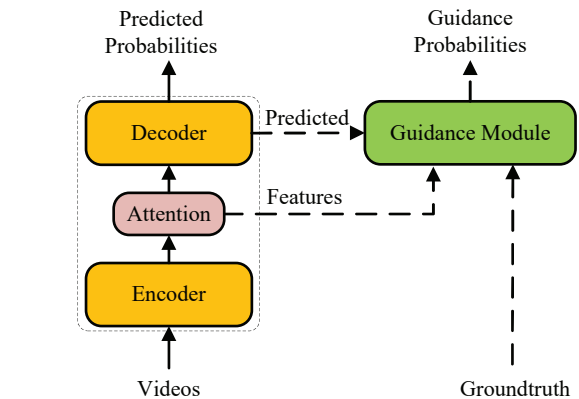}  
	\caption{The overall structure of the proposed GMNet, encoder-decoder framework with guidance module, in which the dotted arrows means that it works when training, but doesn't work when testing.}   
\end{figure}

\section{Related Work}

At present, effective methods to finish video captioning tasks can be divided into two categories: 1) template-based methods; 2) the methods based on neural network. In this section, we will introduce the previous work and development of these two approaches respectively.

Early approaches for video captioning are based on templates. The template is defined first, and the subject, predicate and object are stuffed into the template to generate sentences by means of detection and the like afterwards. In \cite{5}, a method is proposed to output a short sentence that summarizes the main activities in a video, such as actors, actions and their objects to describe a short video clip. A holistic data-driven technique is presented to generate natural language descriptions of videos in \cite{1}. However, the template-based approach leads to rigid and inflexible sentences because templates are defined in advance.

With the rapid development of deep learning, more and more excellent networks \cite{resnet, Inceptionv4} for image feature extraction are proposed and they can be employed as encoders. In \cite{N5}, long short-term memory (LSTM) is adopted as a decoder and a joint embedding model is proposed to explore the relationship between visual content and sentence semantics. However, the above mentioned approache applies pooling technic after extracting video features, which ignores or even loses some significant information at different times in the video. To take into account both the local and global temporal structure of videos to produce better descriptions, in \cite{N8}, the authors propose a time attention mechanism that can transcend local time modeling and learn to automatically select the most relevant time period in the case of text generation. Practice confirms that the performance of using attention mechanism is significantly better than that of using pooling strategy. In \cite{N2}, the authors present an attention-based LSTM model with semantic consistency, considering the attention mechanism that allows the selection of distinctive features of video and the correlation between sentence semantics and visual content. In \cite{N3}, a layered video caption generation method with adjusted time attention is proposed to compensate for the defect that applying attention mechanism to non-visual text will mislead and reduce the overall performance of video captioning. In \cite{N4}, a multimodal random recursive neural network is proposed, which can utilize potential random variables to model the observed uncertainties in the data.

Although the above presented approaches can achieve good results in the video captioning task, as a framework based on encoder-decoder, it is inevitable to adopted teachers' forced strategies to train the network. The prediction probability of each word is close to the 0-1 distribution by training, but the network ignores other words while predicting the word of the moment. This paper proposes a guidance module to combine the context of words generated and future groundtruth at each time step to encourage encoder-decoder framework to generate words that are relevant to the past and future translations. In addition, we discover that normalizing the output of encoder and attention module will greatly improve the performance of encoder-decoder framework in the video captioning task. Our experiments with soft-attention-LSTM \cite{N8} indicate that our presented guidance module and our discovery do improve the translation performance in the video captioning task.

\section{Architecture}
We propose a novel guidance module and add normalization to the encoder-decoder framework. We call our overall network GMNet for video captioning. On the basis of encoder-decoder framework, we find that normalization for input features can improve the performance for video captioning and our proposed guidance module can be capable of compensating for the disadvantage of teachers' forced training focusing only on the current step word. Join the presented guidance module to the encoder-decoder framework in the training process to guide the network generating word associated with the context. During the test, the guidance module is removed and video captioning is performed only by using the weights of the encoder-decoder framework obtained through training. In this paper, we serve soft-attention-LSTM \cite{N8} as the baseline for video captioning. Our proposed GMNet is shown in Figure 1 and the structure is described as follows.

\subsection{Attention-Based Encoder-Decoder}
Encoder-decoder models are extensively applied in NLP and NMT. In video captioning tasks, the purpose is to acquire a sentence $\mathbf{S}=\left\{\mathbf{s}_{1}, \mathbf{s}_{2}, \ldots, \mathbf{s}_{n}\right\}$ to describe the content according to a given video clip $\mathbf{V}$. Classical encoder-decoder architectures model the captioning generation probability directly word by word:

\begin{equation}
P(\mathbf{S} \mid \mathbf{V})=\prod_{i=1}^{n} P\left(\mathbf{s}_{i} \mid \mathbf{s}_{<i}, \mathbf{V} \right)
\end{equation}

Where $n$ denotes the length of the sentence and $\mathbf{s}_{<i}$ represents the already generated partial caption. We apply the CNN encoder to encode the video clip into a specific length feature vector and employ the RNN decoder to generate the corresponding caption.

\subsubsection{Encoder}
Different from the NLP, visual features need to be extracted to capture the high-level semantic information about the video and the CNN is relied on as the encoder. In recent years, a number of excellent networks have emerged to extract image features with the high-level semantic information, such as Resnet \cite{resnet} and InceptionV4 \cite{Inceptionv4}. For the sake of better performance in video captioning, a deep network InceptionV4 is applied to extract the high-level semantic features of the video. We feed frames into InceptionV4 to obtain the features that represent the video $\mathbf{V}$ and we take a fixed number of frame features $\left\{\mathbf{v}_{1}, \mathbf{v}_{2}, \ldots, \mathbf{v}_{m}\right\}$ to train the encoder-decoder, where $m$ is the fixed number.

\subsubsection{Decoder}
The decoder applied to dispose sequence-to-sequence questions is usually an RNN like network and LSTM is widely accepted for similar problems. Compared to the traditional RNN, the advantage of the LSTM is that it could remember and utilize all the previous information to predict the next step, instead of only relying on the neighbouring previous information to predict the next step. A description of the video will be generated by feeding the video features of the encoder into the decoder.

\subsubsection{Attention}
The purpose to apply a soft attention mechanism \cite{attention} is to exploit the temporal ordering of objects and actions across the entire video clip and avoid conflating temporally disparate events. The attention mechanism can extract a part of each frame feature through a weight in each time step to form a new suitable feature. The attention module will get the most appropriate input characteristics to the decoder at each time step, which can improve the performance of the decoder. 

\begin{figure*}  
	\centering  
	\includegraphics[scale=0.8]{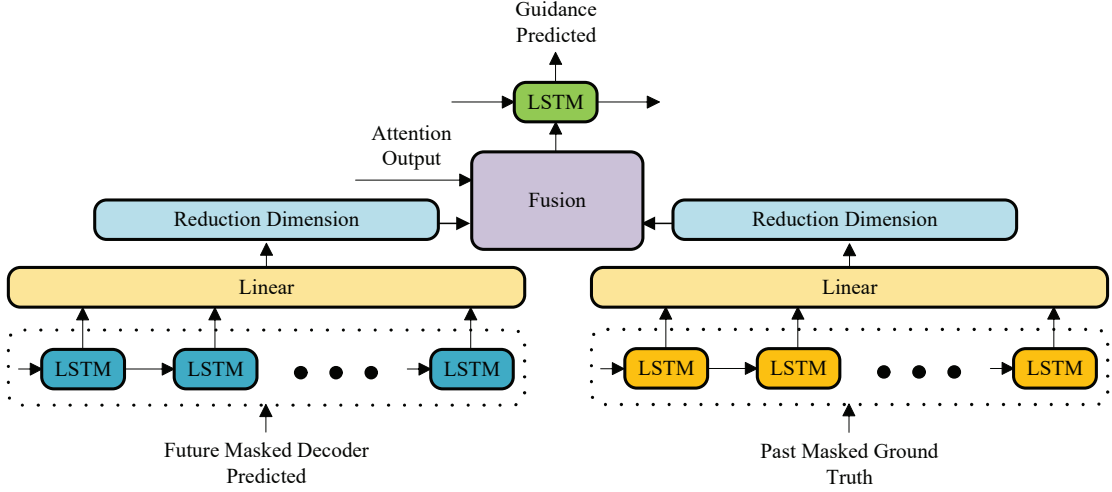}  
	\caption{The structure of the guidance module, we show here a time step of how the guidance module works.}   
\end{figure*}

\subsection{Innovative points}
\subsubsection{Layer normalization}
Before training, the performance can be enhanced by normalizing the data to make the distribution of data consistent. In the process of deep neural network training, batch normalizaiton (BN) \cite{BN} is usually applied for each batch sent into the network. During the training process, the data distribution will change, which will bring difficulties to the learning of the next layer network. The purpose of normalization is that it forces the data back to a normal distribution with mean of 0 and variance of 1, so that the net not only has the same data distribution but also avoids the vanishing gradient. It is convenient to utilize BN for a fixed depth forward neural network, for example, CNN, but for RNN, the length is not consistent. Therefore, layer normalizaiton (LN) \cite{LN} comes into being. The input of neurons in the same layer in layer normaliztion has the same mean value and variance, and different input samples have different mean value and variance, which does not depend on the size of batch and the depth of input sequence. Layer normaliztion calculates mean and variance as follows:

\begin{equation}
\begin{aligned}
& \mu = \frac{1}{H} \sum_{i=1}^{H} a_i\\
& \sigma = \sqrt{\frac{1}{H} \sum_{i=1}^{H} (a_i - \mu)^2}
\end{aligned}
\end{equation}
Where $H$ is the number of neurons in a layer. The network in the same layer shares a mean and variance, and different training samples correspond to different mean and variance.

Although it has been widely accepted that normalization of features in neural networks can improve network performance, surprisingly, no article in video captioning task explicitly proposes the use of normalization of features. We attempt to apply normalization in the proposed GMNet and draw the conclusion that normalization can indeed improve the performance of video captioning. We adopt layer normalization after both the output of encoder and the output of attention module.

\subsubsection{Guidance module}
The detail of our guidance module is shown in Figure 2. In order to combine the past and future words in the caption, at each time step, we input the decoded results and the groundtruth which is masked the current and past time steps into two LSTM networks respectively. The linear transformation of the output results and the features obtained after the attention module are combined and input into an LSTM network which is the same as the decoder. The proposed encoder-decoder-guidance architecture can be trained in an end-to-end fashion. With such a guidance process, the decoder is encouraged to generate a word related to the context at each time step, which is expected to improve the video captioning performance. In practice, the distribution drawn by the guidance module is applied to produce an additional loss to guide the distribution drawn by the encoder-decoder framework. At the test phase, our proposed GMNet leaves out the guidance module and performs inference only with the encoder-decoder framework.

In the time step $i$, we regard the decoded words $\{y_1,...,y_{i-1}\}$ as the past words of caption. Under the teacher's forced strategy, the model will go forward to the sequence, groundtruth $\{y_{i+1}^*,...,y_I^*\}$ after this time step, we consider $\{y_{i+1}^*,...,y_I^*\}$ as the future words of caption.

Given to get a representation of past and future caption, we use two same LSTM as the encoders and fuse the output of the two encoders together. Assume the hidden state matrices outputted by the past and future encoders are $\mathbf{A}_{p}$ and $\mathbf{A}_{f}$ respectively, then the two outputs are fused together as:
\begin{equation}
\mathbf{A}_{e}=RD(\mathbf{W}_{p} \mathbf{A}_{p})+RD(\mathbf{W}_{f} \mathbf{A}_{f})
\end{equation}
Where $\mathbf{A}_{e}$ is the fused vector, $\mathbf{W}_{p}$ and $\mathbf{W}_{f}$ are linear transformations, $RD$ means reduction dimension that sums a two-dimensional matrix in one dimension.
To make the words generated by the current time step reflect the meaning of the appropriate source video, we conducted a simple fusion of the previously fused output $\mathbf{A}_{e}$ and the output of the attention module $\mathbf{A}_{att}$ in the encoder-decoder framework to get a finally feature $\mathbf{A}_{F}$:
\begin{equation}
\mathbf{A}_{F}=Norm(Norm(\mathbf{A}_{e})+ Norm(\mathbf{A}_{att}))
\end{equation}
Where $Norm$ is layer normalization. Finally, we apply an LSTM network as the decoder of our guidance module to decode the finally feature $\mathbf{A}_{F}$ obtained in the previous step to generate the guiding caption.

\subsection{Training}
We conducted joint training on the original decoding results and the guidance results. Specifically, for the original decoding module, a cross-entropy loss is employed as:
\begin{equation}
\begin{aligned}
L = -\sum_{t=1}^{T}\log p(w_t\mid w_{<t},\textbf{x})
\end{aligned}
\end{equation}
The guidance module is also optimized via a cross-entropy loss as:
\begin{equation}
\begin{aligned}
L_e = -\sum_{t=1}^{T}\log p_e(w_t\mid w_{<t},\textbf{x})
\end{aligned}
\end{equation}
Where $L$ is the loss of original decoder, $L_e$ is the loss of guidance module. Train the network with $L_{all}$ as a total loss to encourage the network to produce context-sensitive smoother captions:
\begin{equation}
\begin{aligned}
L_{all} = L + L_e
\end{aligned}
\end{equation}

\section{Experimental Results}

\subsection{Common Dataset MSVD}
We conduct experiments on widely-used dataset MSVD \cite{MSVD} to evaluate the effectiveness of the proposed GMNet. MSVD is microsoft video description corpus which is a famous benchmark for video captioning. It includes 1970 video clips from YouTube, which contains a lot of themes and is very suitable for training and evaluating model for video captioning. We adopt the standard partition method as \cite{4} that devides 1200 videos to form the training set, 100 videos for validation and 670 videos for testing.
\subsection{Evaluation Metrics}
The evaluation indicators used in the video captioning field are based on natural language processing field, which generally include BLEU \cite{BLEU}, METEOR \cite{Meteor}, ROUGE\_L \cite{ROUGE} and CIDEr \cite{CIDEr}. BLEU is the first machine translation evaluation index proposed to compare the n-gram degree of coincidence between the candidate translation and the reference translation. The higher the degree of coincidence is, the higher the quality of the translation is. About METEOR, the synonym set has been expanded with wordnet and other knowledge sources, and the morpheme of words has been considered. ROUGE\_L calculates the longest common subsequence length of the candidate captions and the reference captions, and the longer the length is, the higher the score is. CIDEr is the combination of BLEU and vector space model. It regards each sentence as a document, and then calculates the cosine angle of the vector TF-IDF, on which it depends to obtain the similarity between the candidate sentence and the reference sentence, and evaluate whether the model has captured the key information.

\subsection{Results}
In this section, we present our experimental results. The results are shown in Tables which show the similar methods' perfomance in terms of BLEU\_4, METEOR, ROUGE\_L and CIDEr and the performance of SA (the original LSTM with soft attention), SA\_LN (SA after the feature normalization) and GMNet (SA\_LN with the proposed guidance module) on MSVD respectively.

Table 1 and Table 2 summarize the results on MSVD dataset. Table 1 shows the comparison of GMNet results with other similar frameworks (aLSTMs \cite{N2}, hLSTMat \cite{N3}). Our results are much superior to others in CIDEr although our results are slightly lower than hLSTMat in BLEU\_4 and METEOR. In Table 2, we show the results of ablation experiments. First, we test SA on MSVD. After that, we conduct layer nomalization of the input features and the attention output and re-validate them on SA. The results on four indicators, BLEU\_4, METEOR, ROUGE\_L and CIDEr, show that the operation of layer nomalization on the feature has greatly improved the performance of the network. We note that the scores of BLEU\_4 and CIDEr have been increased by 0.9\% and 1.6\% respectively, and the scores of METEOR and ROUGE\_L have been increased by 0.2\% and 0.5\% respectively. Next, we add the proposed guidance module into the network. Compared with SA\_LN, we notice that the score of BLEU\_4 and CIDEr are increased by 0.8\% and 1.4\% respectively, the score of ROUGE\_L are improved by 0.4\%, but the score of METEOR does not change. The above results verify our hypothesis. The improved scores of BLEU\_4 and CIDEr prove that our guidance module could improve the accuracy and fluency of the generated captions to some extent. 

\begin{table}[]
	\caption{Performance evaluation of different video captioning models on the MSVD dataset in terms of BLEU\_4, METEOR, ROUGE\_L, and CIDEr scores (\%)}
	\centering
	\begin{tabular}{c|c|c|c|c}
\hline
		Model                     & BLEU\_4       & METEOR        & ROUGE\_L      & CIDEr         \\ \hline
		
		aLSTMs\cite{N2}                    & 50.8          & 33.3          & -             & 74.8          \\ \hline
		hLSTMat\cite{N3}                  & \textbf{53.0}          & \textbf{33.6}          & -             & 73.8          \\ \hline
		GMNet (ours) & 52.1 & 33.5          & - & \textbf{83.1} \\ \hline
	\end{tabular}
\end{table}

\begin{figure}[]  
	\centering  
	\includegraphics[scale=1]{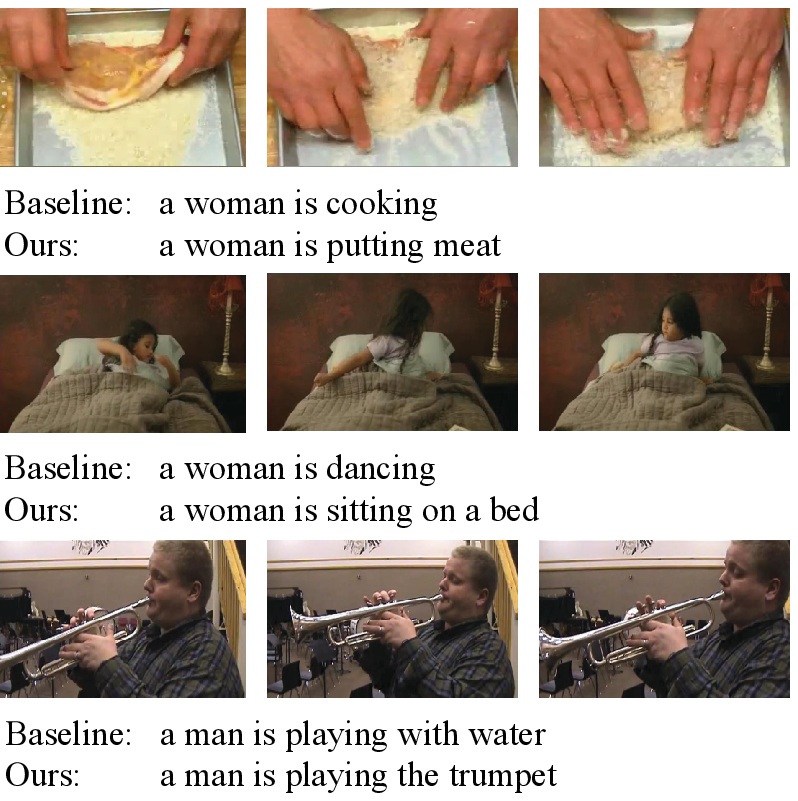}  
	\caption{Visualization of some video captioning examples on the MSVD dataset with different models. "Baseline" denotes captions generated by LSTM with soft attention; "Ours" denotes captions generated by our proposed GMNet.
	}   
\end{figure}

\begin{table}[H]
	\caption{Performance evaluation of SA, SA\_LN and GMNet on the MSVD dataset in terms of BLEU\_4, METEOR, ROUGE\_L, and CIDEr scores (\%)}
	\centering
	\begin{tabular}{c|c|c|c|c}
		\hline
		Model               & BLEU\_4       & METEOR        & ROUGE\_L      & CIDEr         \\ \hline
		SA (ours)            & 50.4          & 33.3          & 69.8          & 80.1          \\ \hline
		SA\_LN (ours)      & 51.3          & \textbf{33.5}          & 70.3 		& 81.7          \\ \hline
		GMNet (ours) & \textbf{52.1} & \textbf{33.5}          & \textbf{70.7} & \textbf{83.1} \\ \hline
	\end{tabular}
\end{table}

It should be emphasized that the method we propose would not be in confrontation with the most advanced method. The worthness of the proposed module would be to improve the performance of the encoder-decoder framework in the task of video captioning. In Figure 3, we show some video captioning examples generated by baseline SA and our proposed GMNet on the MSVD dataset respectively. We notice that captions generated by GMNet are more accurate and smoother than captions generated by the baseline.

\section{Conclusion}
In this paper, we show the finding that layer nomalization of encoder output and attention output can improve the performance of encoder-decoder framework for video captioning task, and put forward a guidance module to compensate for a drawback that encoder-decoder framework's application of teachers' mandatory strategy to train. By combining the captions generated by the decoder with the groundtruth captions, our proposed GMNet is realized. It is necessary to emphasize that our method is not superior to the most advanced method, but that our method can be applied to any encoder-decoder framework. Experiments on MSVD show that the proposed method does improve the performance of baseline for video captioning.

\bibliographystyle{plain}
\bibliography{bibfile}

\end{document}